\pdfoutput=1

\documentclass[11pt]{article}
\usepackage[utf8]{inputenc} 
\usepackage{xcolor}
\usepackage{tabularray}
\usepackage{siunitx}
\usepackage{CJKutf8}
\usepackage{siunitx}
\usepackage{float}    
\usepackage{CJK}
\usepackage[final]{acl}
\usepackage{CJKutf8}
\usepackage{graphicx}
\usepackage{booktabs}
\usepackage{bm}
\usepackage{tikz}
\usetikzlibrary{bayesnet}
\usepackage{times}
\usepackage{blindtext}
\usepackage{microtype}
\usepackage{latexsym}
\usepackage{amsmath}
\usepackage{makecell}
\usepackage{amssymb}
\usepackage{dsfont}
\usepackage{graphicx}
\usepackage{subcaption}
\usepackage{multirow}
\usepackage{colortbl}
\usepackage[table]{xcolor}
\usepackage{booktabs}
\usepackage{threeparttable}
\usepackage{algorithm}
\usepackage{algorithmic}
\usepackage{float}
\usepackage{siunitx}
\usepackage{siunitx}
\usepackage{siunitx}
\usepackage{enumitem} 

\usepackage[T1]{fontenc}

\usepackage[utf8]{inputenc}

\usepackage{microtype}

\usepackage{inconsolata}

\usepackage{graphicx}

\usepackage{multirow}
\usepackage[normalem]{ulem}
\useunder{\uline}{\ul}{}

\title{XTRA: Cross-Lingual Topic Modeling with Topic and Representation Alignments}

\author{
 \textbf{Tien Phat Nguyen\textsuperscript{1}\footnotemark[1]},
 \textbf{Vu Minh Ngo\textsuperscript{1}\footnotemark[1]},
 \textbf{Tung Nguyen\textsuperscript{1}},
 \textbf{Linh Van Ngo\textsuperscript{1}\footnotemark[2]},
\\
 \textbf{Duc Anh Nguyen\textsuperscript{1}},
 \textbf{Sang Dinh\textsuperscript{1}},
 \textbf{Trung Le\textsuperscript{2}},
\\
 \textsuperscript{1} Hanoi University of Science and Technology, Vietnam,\\
 \textsuperscript{2} University of Monash, Australia,
}

\begin{document}
\maketitle

\renewcommand{\thefootnote}{*} 
\renewcommand{\thefootnote}{$\dagger$} 

\footnotetext[1]{Equally contributed.} 
\footnotetext[2]{Corresponding author: \href{mailto:email@domain}{linhnv@soict.hust.edu.vn}} 
\renewcommand*{\thefootnote}{\arabic{footnote}}


\begin{abstract} 
    
    Cross-lingual topic modeling aims to uncover shared semantic themes across languages. Several methods have been proposed to address this problem, leveraging both traditional and neural approaches. While previous methods have achieved some improvements in topic diversity, they often struggle to ensure high topic coherence and consistent alignment across languages.  We propose \textbf{XTRA} (\textbf{Cross}-Lingual Topic Modeling with \textbf{T}opic and \textbf{R}epresentation \textbf{A}lignments), a novel framework that unifies Bag-of-Words modeling with multilingual embeddings. XTRA introduces two core components: (1) representation alignment, aligning document-topic distributions via contrastive learning in a shared semantic space; and (2) topic alignment, projecting topic-word distributions into the same space to enforce cross-lingual consistency. This dual mechanism enables XTRA to learn topics that are interpretable (coherent and diverse) and well-aligned across languages. Experiments on multilingual corpora confirm that XTRA significantly outperforms strong baselines in topic coherence, diversity, and alignment quality. Code and reproducible scripts are available at \url{https://github.com/tienphat140205/XTRA}.

\end{abstract}

\section{Introduction}\label{section:introduction}

\begin{figure}
        \centering
        \includegraphics[width=0.48\textwidth]{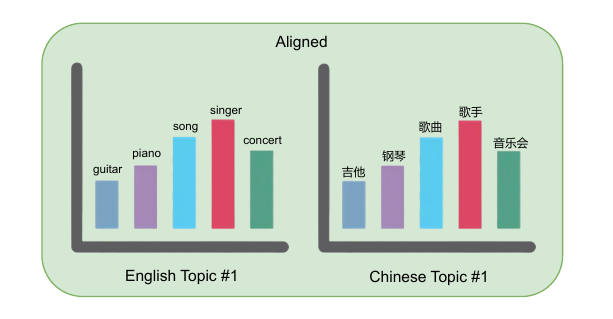}
        \caption{An example of cross-lingual topic alignment: both English and Chinese word clusters describe the shared theme of music.}
        \label{fig:figure1}
    \end{figure}

    Identifying latent thematic structures within large text corpora is a central goal in computational linguistics, with topic modeling (TM) serving as a foundational technique \cite{1999plsi, blei2003lda}. Extending this capability to the multilingual setting has led to the development of Cross-Lingual Topic Modeling (CLTM), which aims to uncover shared latent topics across languages despite lexical and structural differences \cite{DBLP:conf/www/NiSHC09, MimnoWNSM09, YuanDY18, WuLZM20,infoctm}. CLTM models are essential for bridging cultural and linguistic divides, providing a means to understand and compare shared narratives as well as distinct global perspectives. As illustrated in Figure~\ref {fig:figure1},  cross-lingual topic models aim to uncover semantically similar themes across languages, such as English and Chinese word clusters that both describe the concept of music. This illustrates how cross-lingual topic models can discover thematic overlap despite significant differences in surface forms.

    Despite significant progress, developing robust and broadly applicable Cross-Lingual Topic Models (CLTMs) continues to pose substantial challenges. Dictionary-based approaches remain attractive among the most common strategies due to their simplicity and interpretability. However, these methods are fundamentally limited by the coverage and quality of bilingual lexical resources, which are often insufficient, especially when dealing with low-resource languages or domain-specific vocabularies. This low-coverage problem has been consistently observed across multiple prior works~\cite{DBLP:conf/acl/ShiLBX16, YuanDY18, WuLZM20}, where the lack of comprehensive term mappings hampers the alignment of topics across languages and reduces the overall semantic fidelity of the model \cite{infoctm}. In addition to lexical sparsity, these CLTMs often suffer from repetitive topic collapse, where multiple topics converge toward semantically similar word distributions~\cite{infoctm}. This redundancy undermines topic diversity and interpretability, particularly in multilingual settings where fine-grained distinctions are critical.

    Advanced neural topic model InfoCTM \cite{infoctm}) addresses redundancy and topic collapse by applying contrastive objectives on the decoder’s topic–word distributions ($\beta$). These methods increase topic diversity and yield more coherent topics within each language. Yet two limitations remain. First, they often rely on language-specific encoders or independent parameterization, which restricts scalability and leaves $\beta$ only loosely aligned across languages. More critically, most cross-lingual topic models \cite{infoctm,WuLZM20} neglect the refinement of document–topic proportions ($\theta$) across languages. This omission is consequential: without cross-lingual consistency at the $\theta$ level, topic mixtures cannot be compared reliably, weakening semantic alignment across languages. As observed in Table~\ref{tab:infoctm_misalignment}, English–Chinese topics may each appear internally coherent yet still fail to correspond semantically, highlighting the limitations of current approaches.

    \setlength{\tabcolsep}{1pt}     
    
    \captionsetup[table]{font=small, skip=2pt}
    
    \begin{CJK}{UTF8}{gbsn}
    
    \begin{table}[t]
    \centering
    \resizebox{\linewidth}{!}{%
    \begin{tabular}{llcccccc}
        \hline
        \textbf{En Topic\#1:} & & shampoo & lotion & fragrance & clay & powder  \\
        \textbf{ZH Topic\#1:} & & 婴儿 & 墨水 & 涂 & 糊 & 干  \\
        \textbf{Translation\#1:}   & & baby & ink & smear & paste & dry  \\
        \hline
        \textbf{En Topic\#2:} & & violin & orchestra & rhythm & performance & classical  \\
        \textbf{ZH Topic\#2:} & & 按摩 & 枕头 & 鞋 & 滑 & 脖子  \\
        \textbf{Translation\#2:}   & & massage & pillow & shoe & slippery & neck  \\
        \hline
        \textbf{En Topic\#3:} & & translation & translator & original & poem & English \\
        \textbf{ZH Topic\#3:} & & 译文 & 文言文 & 原文 & 思录  & 改动 \\
        \textbf{Translation\#3:}   & & T. Trans & C. Chinese & Ori. Text & reflection   & revision \\
        \hline
    \end{tabular}
    }
    \caption{Example of misaligned topics generated by InfoCTM across English and Chinese. The words grouped under each topic differ in semantic coherence between the two languages.}
    \label{tab:infoctm_misalignment}
    \vspace{-1em}  
    \end{table}
    
    \end{CJK}

To address these challenges, we propose \textbf{XTRA} (Cross-Lingual Topic Modeling with Topic and Representation Alignments), a unified framework designed for robust cross-lingual topic discovery that moves beyond complex language-specific encoders. Instead, XTRA utilizes lightweight MLPs to project Bag-of-Words (BoW) inputs into a shared space, where a common encoder effectively captures both semantic structure and crucial cross-lingual alignment signals. One of the key ideas of our model is a clustering-guided contrastive learning objective that refines document-topic distributions. By clustering documents from different languages into semantically coherent groups, XTRA identifies latent cross-lingual themes. A contrastive loss then pulls together documents with similar thematic content irrespective of language while pushing apart those with dissimilar themes, fostering topic distributions that are both discriminative and consistent across languages.
    
    In parallel, XTRA introduces a novel approach to align topic-word distributions across languages. Instead of relying on vocabulary matching or dictionary lookups, we learn to project topic-word distributions into a shared semantic space using trainable transformation layers. In this space, a second contrastive objective brings semantically equivalent topics across languages closer together. This dual-contrastive design allows XTRA to learn both high-quality document-topic structures and semantically aligned topic-word meanings without any parallel supervision. Our main contributions are:
    \begin{itemize}
        \item We propose \textbf{XTRA}, a contrastive cross-lingual topic modeling framework that directly leverages powerful multilingual contextual embeddings via a shared encoder.
        \item We design a clustering-based contrastive learning strategy that improves document-topic distributions by aligning documents sharing similar semantics across languages.
        \item We develop a semantic-space alignment technique for topic-word distributions using projection and contrastive loss, enabling robust cross-lingual topic equivalence.
        \item We conduct extensive experiments on multilingual datasets, showing that XTRA surpasses state-of-the-art CLTM models in topic coherence, topic diversity, and cross-lingual alignment.
    \end{itemize}

\begin{figure*}
\centering
\includegraphics[width=1\textwidth]{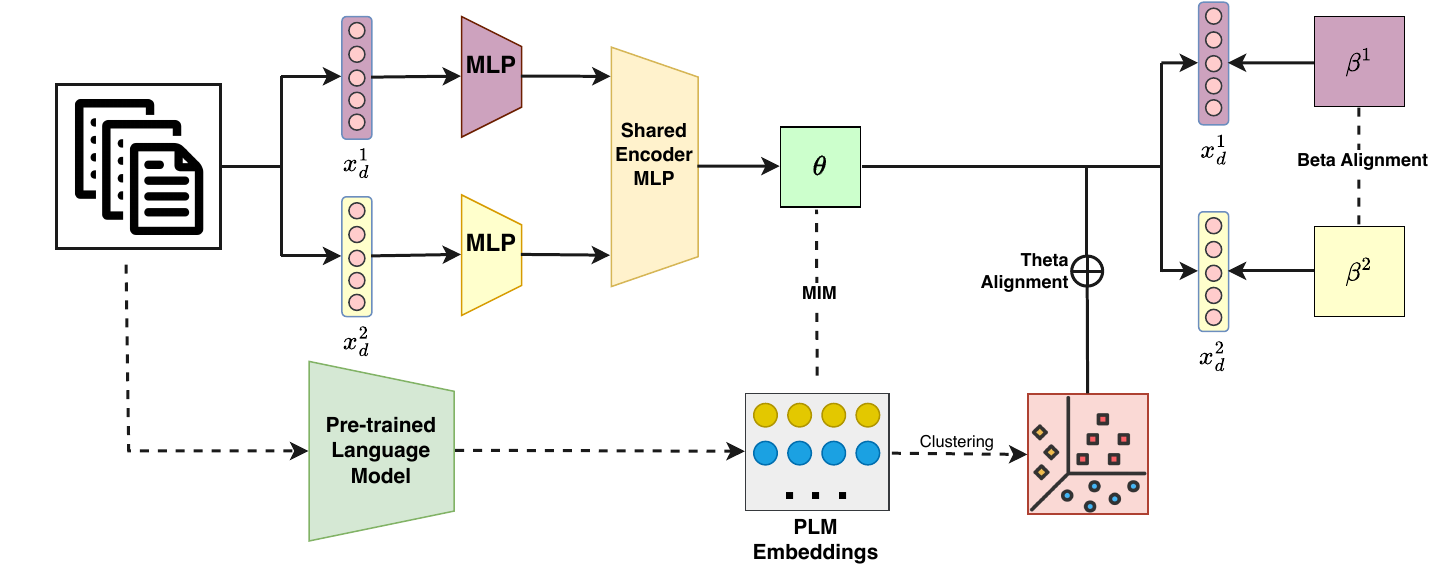}
\caption{The overall architecture of XTRA. Our method processes language-specific Bag-of-Words inputs ($x_d^1$, $x_d^2$) through dedicated MLPs into a Shared Encoder to estimate document-topic distributions ($\theta$). XTRA proposes a novel dual-alignment approach using Pre-trained Language Model (PLM) embeddings: aligning document-topic distributions ($\theta$) via MIM and clustering (Theta Alignment), and aligning topic-word distributions ($\beta^1$, $\beta^2$) semantically (Beta Alignment) to achieve cross-lingual consistency.}
\label{fig:model}
\end{figure*}

\section{Preliminaries} \label{sec:preliminaries}

\subsection{Notations} \label{subsec:notations}

 We model a multilingual corpus comprising $D$ documents in two languages, denoted $L_1$ and $L_2$, with the objective of discovering $K$ shared topics. The corpus is represented as a collection $X = \{x_d\}_{d=1}^D$ of Bag-of-Words (BoW) representations, where each document $x_d$ belongs to either $L_1$ or $L_2$. The vocabulary for $L_1$ is $V_1$ of size $|V_1|$, and for $L_2$ is $V_2$ of size $|V_2|$. The BoW representation of document $d$ is $x_d \in \mathbb{R}^{|V_\ell|}$, where $\ell \in \{1, 2\}$ indicates the document’s language. A pre-trained multilingual language model (MLM) provides embeddings $x_{d\text{PLM}} \in \mathbb{R}^M$ for each document $d$, where $M$ is the MLM embedding dimension. Applying a clustering algorithm to $\{x_{d\text{PLM}}\}_{d=1}^D$ produces $K$ clusters, grouping documents by semantic similarity across languages.

For each language $L_{\ell}$ where $\ell\in \{1, 2\}$, the topic-word distribution is denoted by $\beta^{(\ell)} \in \mathbb{R}^{|V_\ell| \times K} = (\beta_1^{(\ell)}, \ldots, \beta_K^{(\ell)})$, where each $\beta_k^{(\ell)} \in \mathbb{R}^{|V_\ell|}$ represents the word distribution for topic $k$ over the vocabulary $V_\ell$, satisfying $\displaystyle\sum_{v \in V_\ell} \beta_{v,k}^{(\ell)} = 1$. Each document $x_d$ is associated with a topic proportion vector $\theta_d \in \mathbb{R}^K$ such that $\displaystyle\sum_{k=1}^K \theta_{d,k} = 1$.

\subsection{VAE-based Topic Model} \label{subsec:VAE-based Topic Model}
Our foundational structure employs a Variational Autoencoder (VAE). The encoder maps a document's Bag-of-Words (BoW) representation $x_d$ to parameters $(\mu, \Sigma)$ of a posterior $q(z|x_d)=\mathcal{N}(z|\mu, \Sigma)$. A latent variable $z$ is sampled (via reparameterization~\cite{kingma2013vae}) from this posterior, with $p(z)=\mathcal{N}(z|\mu_0, \Sigma_0)$ as the prior. The topic proportion is then $\theta_d=\mathrm{softmax}(z)$. The decoder models topic-word distributions $\beta \in \mathbb{R}^{V\times K}$ (e.g., inferred via optimization~\cite{srivastava2017prodlda} or learned embeddings ~\cite{dieng2020etm,ecrtm}) and reconstructs $x_d$ by sampling from $\mathrm{Multinomial}(\mathrm{softmax}(\beta\theta_d))$. The training objective is:
\begin{equation*}
    \begin{split}
    \mathcal{L}_{\mathrm{TM}} = \frac{1}{D} \sum_{d=1}^{D} \Big[& - (x_d)^{\top} \log (\mathrm{softmax}(\beta \theta_d)) \\ &+ \mathrm{KL}(q(z \vert x_d) \| p(z)) \Big]
    \end{split}
\end{equation*}

\section{Methodology}\label{sec:method}

    We propose XTRA, a novel cross-lingual topic modeling framework whose overall architecture is illustrated in Figure~\ref{fig:model}. XTRA introduces an innovative inference mechanism for multilingual documents and enhances topic quality through contrastive alignment between topic-word distributions and document-topic distributions. Additionally, we leverage contextualized multilingual embeddings to capture nuanced semantic representations and gain a deeper understanding of the corpus content.

\subsection{Enhanced Cross-Lingual Encoder}
\label{subsec:3_1}
The encoder transforms document representations into latent topic distributions $\theta$. Prior cross-lingual topic models (e.g., \cite{infoctm}) often use separate encoders for each language, increasing the number of parameters and potentially hindering inherent cross-lingual $\theta$ alignment, thus requiring complex post-processing.

    We propose an enhanced encoder that uses a single, shared core to process language-specific inputs. Language-specific Bag-of-Words (BoW) inputs ($x_{d}^\ell$) are first processed by language-specific Multi-Layer Perceptrons (MLPs), which project BoW vectors into a common, language-agnostic space before entering the shared encoder. This reduces parameters and encourages the shared encoder to learn a latent space where $\theta$ exhibits better intrinsic cross-lingual alignment.

    To further improve $\theta$'s quality and cross-lingual consistency, we leverage large multilingual embeddings~\cite{bgem3}. Inspired by \citet{pham2024neuromaxenhancingneuraltopic}, a contrastive objective aligns $\theta$ with this shared semantic space.
    
    This guidance uses an InfoNCE loss \cite{oord2019infonce}:
    \[
I(\mathbf{X}_{\text{PLM}}; \Theta) \geq \log B + \mathcal{L}_{\text{InfoNCE}}
\]
\begin{equation*}
\resizebox{0.99\linewidth}{!}{$
    \mathcal{L}_{\text{InfoNCE}} = - \dfrac{1}{D} \displaystyle\sum_{i=1}^{D}
    \log 
        \dfrac{
            \exp(f(\theta_i, x_{i\text{PLM}}))
        }{
            \sum_{\theta' \in B_i} \exp(f(\theta', x_{i\text{PLM}}))
        }
$}
\end{equation*}

    $B_i$ holds sampled topic proportions for document $i$ (positive/negative for $x_{i\text{PLM}}$), drawn from the same batch as $x_i$ with constant size $B$. $f(\theta, x_{\text{PLM}}) = \dfrac{\langle\phi_\theta(\theta), x_{\text{PLM}}\rangle}{\|\phi_\theta(\theta)\|\|x_{\text{PLM}}\|}$ (using learnable $\phi_\theta$) measures similarity between $\theta$ and $x_{\text{PLM}}$. Minimizing this loss improves $\theta$'s semantic relevance and cross-lingual alignment by encouraging similarity with corresponding embeddings.

\subsection{Topic Distribution Clustering Contrastive Alignment}
\label{subsec:topic_distribution_clustering_contrastive_alignment_academic}

Aligning document-topic distributions ($\theta$) across languages is crucial but has often been overlooked. We introduce Topic Distribution Clustering Contrastive Alignment, a novel mechanism that explicitly shapes the $\theta$ space through cross-lingual clustering and contrastive objectives.

This method achieves cross-lingual $\theta$ alignment using document relationships from 
prior clustering (Figure~\ref{fig:figure2}). Documents are clustered via multilingual 
embeddings \cite{bgem3} (see Appendix~\ref{appendix:clustering} for details), identifying related documents across languages without requiring direct 
translations or parallel data. For a given document’s topic distribution, those from the same cluster act as positive examples, while those from different clusters serve as negative ones. The contrastive objective encourages the document’s representation to be close to the positive examples and distant from the negative ones.

This direct contrastive pressure on $\theta$ steers the encoder towards a latent $\theta$ 
space where proximity reflects cluster-defined cross-lingual similarity, ensuring 
similar-themed documents across languages map to similar topic distribution points. 
This yields more robust, coherent, and aligned document representations. This $\theta$ 
alignment also improves topic interpretability (similar documents load onto similar 
topic profiles) and aids learning sharper, coherent topic-word distributions ($\beta$) 
via a cleaner decoder signal.

This direct $\theta$ alignment is formalized using the multi-positive InfoNCE loss \cite{oord2019infonce}:

    \vspace{-1em}
\begin{equation*}
\scalebox{1.0}{$
    \mathcal{L}_{\text{Cluster}} = -\dfrac{1}{D} \displaystyle\sum_{i=1}^{D} \sum_{\theta_j \in B^+_i} \log S_{ij}
$}
\end{equation*}

Which $S_{ij} = \dfrac{\exp (g(\theta_i, \theta_j))}{\Sigma_{\theta'\in B_i}\exp (g(\theta_i, \theta'))}$, $B_i$ denote the set of sampled topic proportions for document $i$, $B_i^+ \subset B_i$ represent the subset of positive samples $\theta_i^+$ that are semantically related to a given topic proportion $\theta$,  and $\text{g}(\cdot,\cdot)$ is a similarity function (e.g., cosine similarity). Minimizing 
$\mathcal{L}_{\text{Cluster}}$ directly enhances the cross-lingual alignment and 
relevance of the learned document-topic distributions $\theta$.

\begin{figure}
        \centering
        \includegraphics[width=0.48\textwidth]{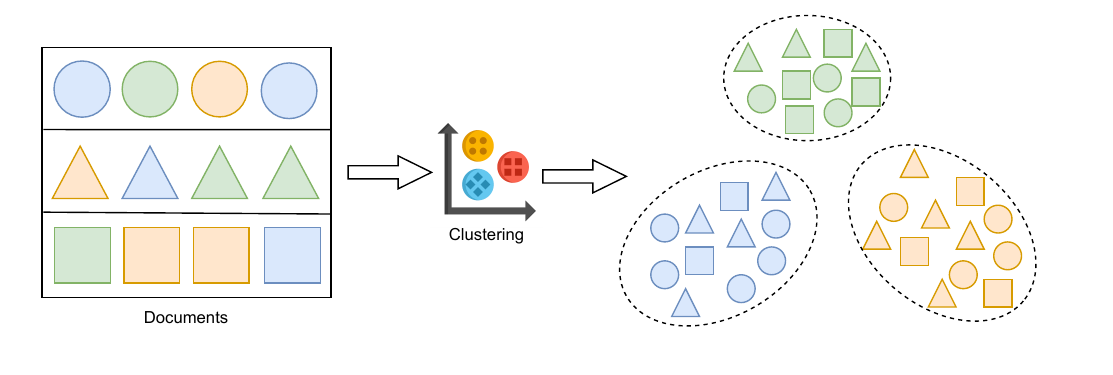}
        \caption{Our clustering-based contrastive alignment illustration. We group similar documents across languages into clusters using multilingual embeddings. Each document is aligned with its cluster via contrastive learning on topic distributions ($\theta$), encouraging cross-lingual consistency in the topic space.}
        \label{fig:figure2}
    \end{figure}
\subsection{Topic-Word Distribution Semantic Alignment}
\label{subsec:topic_word_distribution_semantic_alignment_academic}

Ensuring semantic equivalence of learned topics across languages, by aligning their 
topic-word distributions ($\beta$), is vital for cross-lingual interpretability. 
However, directly aligning these high-dimensional topic distributions poses significant challenges, as it requires effective transformations to handle the complexity and discrepancies between languages.

We propose transferring vocabulary item representations from high-dimensional 
vocabulary space to a lower-dimensional semantic space for alignment. For each 
language $\ell \in \{1,2\}$, a learnable language-specific projection $P_\ell$ 
(typically an MLP) maps a topic's word distribution representation $\beta_k^{(\ell)}$ 
to a shared $D_{\text{sem}}$-dimensional semantic space. Given $\beta^{(1)}, \beta^{(2)}$, 
these projections yield $K$ semantic vectors per language, where $y_k^{(\ell)}$ is topic 
$k$'s projected semantic profile in language $\ell$.

Alignment is enforced in this shared semantic space via an InfoNCE-based contrastive 
objective\cite{oord2019infonce} on projected topic vectors. This maximizes similarity between corresponding 
topic vectors ($y_k^{(1)}, y_k^{(2)}$) relative to non-corresponding ones 
($y_{k'}^{(2)}$), directly aligning their semantic representations. This contrastive 
learning on projected topic-word distributions also fosters greater topic distinction 
by pushing non-corresponding topic vectors apart, potentially reducing overlap and 
improving Topic Uniqueness (TU).

The contrastive loss for $\beta$ alignment is formalized as:
\begin{equation*}
\mathcal{L}^{(1 \to 2)} = - \frac{1}{K} \sum_{k=1}^{K} \log \frac{\exp(g(y^{(1)}_k, y^{(2)}_k))}{\sum_{k'=1}^{K} \exp(g(y^{(1)}_k, y^{(2)}_{k'}))}
\label{eq:l_beta_12}
\end{equation*}
\begin{equation*}
\mathcal{L}^{(2 \to 1)} = - \frac{1}{K} \sum_{k=1}^{K} \log \frac{\exp(g(y^{(2)}_k, y^{(1)}_k))}{\sum_{k'=1}^{K} \exp(g(y^{(2)}_k, y^{(1)}_{k'}))}
\label{eq:l_beta_21}
\end{equation*}
\begin{equation*}
\mathcal{L}_{\beta} = \frac{1}{2} (\mathcal{L}^{(1 \to 2)} + \mathcal{L}^{(2 \to 1)})
\label{eq:l_beta_combined}
\end{equation*}
where $y_k^{(\ell)}$ is the projected semantic vector for topic $k$ in language $\ell$, 
and $g(\cdot, \cdot)$ is a similarity function in 
the projected space. This loss directly compels the projection functions and the 
underlying topic representations to align semantically corresponding topics across 
languages in the shared $D_{\text{sem}}$-dimensional space, facilitating interpretable 
cross-lingual topics.

\begin{table*}[!htbp]
\centering
\small
\setlength{\tabcolsep}{1.5mm}
\renewcommand{\arraystretch}{1}
\begin{tabular}{l ccc ccc ccc}
\toprule
\multirow{2}{*}{Model}
& \multicolumn{3}{c}{EC News}
& \multicolumn{3}{c}{Amazon Review}
& \multicolumn{3}{c}{Rakuten Amazon} \\
\cmidrule(lr){2-4} \cmidrule(lr){5-7} \cmidrule(lr){8-10}
& CNPMI & TU & TQ & CNPMI & TU & TQ & CNPMI & TU & TQ \\
\midrule
MCTA\(^{\dagger}\)      & 0.025  & 0.489  & 0.012  & 0.028 & 0.319 & 0.009 & 0.021 & 0.272 & 0.006 \\
MTAnchor\(^{\dagger}\)  & -0.013 & 0.192  & 0.000  & 0.028 & 0.323 & 0.009 & -0.001 & 0.214 & 0.000 \\
NMTM\(^{\dagger}\)      & 0.031  & 0.784  & 0.024  & 0.042 & 0.732 & 0.031 & 0.009  & 0.679 & 0.006 \\
InfoCTM\(^{\dagger}\)   & 0.048  & 0.913  & 0.044  & 0.043 & 0.923 & 0.040 & 0.034  & 0.870 & 0.030 \\
u\text{-}SVD            & \textbf{0.083}  & 0.830  & 0.069  & 0.054 & 0.638 & 0.034 & 0.025 & 0.584 & 0.015 \\
SVD\text{-}LR           & 0.081  & 0.827  & 0.067  & 0.053 & 0.631 & 0.033 & 0.026 & 0.567 & 0.015 \\
\midrule
\textbf{XTRA}
& 0.076 & \textbf{0.993} & \textbf{0.075}
& \textbf{0.055} & \textbf{0.980} & \textbf{0.054}
& \textbf{0.035} & \textbf{0.975} & \textbf{0.034} \\
\bottomrule
\end{tabular}
\caption{Comparison of CNPMI, TU, and TQ across datasets, where \( \mathrm{TQ}=\max(0,\mathrm{CNPMI})\times \mathrm{TU} \); the best value in each column is in \textbf{bold}, and \(^{\dagger}\) denotes results reported in \cite{infoctm}.}
\label{tab:infoctm}
\end{table*}

\subsection{Overall Objective}\label{sec:overall}
    Inspired by prior work \cite{tu,prior2} suggesting prior modification for better semantic structure capture, we adopt a concise cluster-based Gaussian prior for latent variable $z$ (from which $\theta$ is derived). For $T$ clusters ($K=T$), document counts $n_k$ per cluster, and concentration parameters $\alpha_k = n_k + \epsilon$, the prior $p(z)$, following \citet{srivastava2017prodlda}, is defined with mean and variance:
\begin{equation*}
\mu_{\text{prior},k} = \log \alpha_k - \frac{1}{T} \sum_{j=1}^{T} \log \alpha_j
\end{equation*}
\begin{equation*}
\sigma_{\text{prior},k}^2 = \frac{1}{\alpha_k} \left(1 - \frac{2}{T}\right) + \frac{1}{T^2} \sum_{j=1}^{T} \frac{1}{\alpha_j}
\end{equation*}
This $p(z)$ replaces the standard unit Gaussian prior in the KL divergence term of $\mathcal{L}_{\text{TM}}$.

The overall objective of XTRA is:
\begin{equation*}
\mathcal{L} = \mathcal{L}_{\text{TM}} + \lambda_1 \mathcal{L}_{\text{InfoNCE}} + \lambda_2 \mathcal{L}_{\text{Cluster}} + \lambda_3 \mathcal{L}_{\beta}
\end{equation*}
    where $\lambda_1, \lambda_2, \lambda_3$ balance terms. Minimizing $\mathcal{L}$ ensures interpretable topics and cross-lingual alignment, leveraging the cluster's ability to effectively reflect corpus content.
Further details of the algorithm can be found in Algorithm~\ref{alg:xtra_glocom_style_condensed}.

\section{Experiments and Results}

\subsubsection*{Datasets}
Our experimental evaluations utilized three benchmark datasets: \textbf{EC News} \cite{WuLZM20}, a collection of English and Chinese news articles spanning six categories (business, education, entertainment, sports, technology, fashion); \textbf{Amazon Review} \cite{YuanDY18}, comprising English and Chinese Amazon reviews adapted for a binary classification task where five-star ratings are labeled “1” and others “0”; and \textbf{Rakuten Amazon}, consisting of Japanese reviews from Rakuten and English reviews from Amazon \cite{YuanDY18}, similarly formulated as a binary task based on ratings.

\subsubsection*{Baseline Models}
We evaluated our proposed method against several established baselines: \textbf{MCTA} \cite{DBLP:conf/acl/ShiLBX16}, a probabilistic framework for cross-lingual topic modeling (CLTM) designed to identify cultural variations; \textbf{MTAnchor} \cite{YuanDY18}, which utilizes multilingual anchor words to establish cross-language connections; \textbf{NMTM} \cite{WuLZM20}, a neural approach aligning multilingual topic representations within a common vocabulary space; and \textbf{InfoCTM} \cite{infoctm}, which employs mutual information maximization, often via contrastive objectives, to enhance cross-lingual topic representation alignment and address topic repetition; as well as two clustering-based refinement baselines, \textbf{u-SVD} and \textbf{SVD-LR} \cite{refining}.
\subsubsection*{Evaluation Metrics}
To comprehensively assess generated topic quality and utility, we employed a multi-faceted strategy. For intrinsic quality, we measured cross-lingual topic coherence using \textbf{CNPMI} (Cross-lingual Normalized Pointwise Mutual Information \cite{cnpmi}), an NPMI~\cite{npmi} extension for alignment assessment, and topic diversity using \textbf{TU} (Topic Uniqueness \cite{tu}) for redundancy evaluation; both used the top 15 words per topic. Following \cite{refining}, we report \textbf{TQ (Topic Quality)}, which combines cross-lingual coherence with uniqueness while clipping negative coherence to zero.
To evaluate practical utility and feature transferability, document-topic distributions were used as features for \textbf{SVM-based classification} in intra-lingual (-I) and cross-lingual (-C) settings, following common practice (e.g., \citet{YuanDY18, infoctm}). 
Additionally, inspired by \citet{llm_eval}, \textbf{LLM-based ratings} (1-3 scale) were utilized to assess {intra-lingual coherence} and {cross-lingual topic alignment}.

\subsection{Topic quality analysis}
We evaluate cross-lingual topic quality via coherence (CNPMI), diversity (Topic Uniqueness, TU), and a composite Topic Quality (TQ). For comparison, we include both older baselines (numbers from \citet{infoctm}) and more modern baselines, u-SVD and SVD-LR, which we re-tune under the same embedding setup for fairness (Table~\ref{tab:infoctm}). On EC News, XTRA records a CNPMI of \(0.076\) versus \(0.083\) for u-SVD, yet it leads on TU \(0.993\) against \(0.830\) and on TQ \(0.075\) against \(0.069\). On Amazon Review, CNPMI reaches \(0.055\) for XTRA, with u-SVD second at \(0.054\) and SVD-LR third at \(0.053\); TU and TQ also favor XTRA at \(0.980\) and \(0.054\) versus the strongest alternative at \(0.923\) and \(0.040\). On Rakuten Amazon, CNPMI is \(0.035\) for XTRA, followed by InfoCTM at \(0.034\) and SVD-LR at \(0.026\); TU and TQ again lead at \(0.975\) and \(0.034\) over the next best at \(0.870\) and \(0.030\). Taken together, XTRA consistently {outperforms} the baselines across datasets and metrics; even when clustering models occasionally yield higher CNPMI, they lack an explicit topic space and cannot infer topic distributions for unseen documents.

\begin{figure*}[!t]
    \centering
    \includegraphics[width=\linewidth]{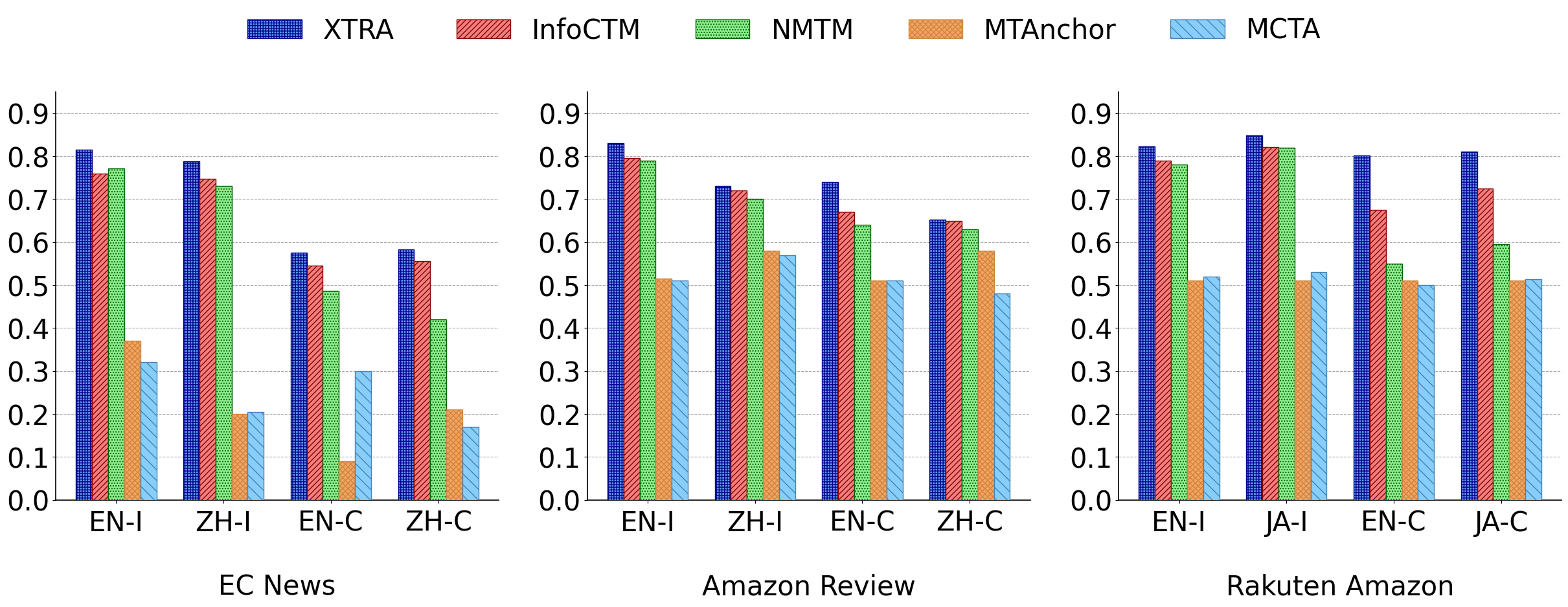}
    \caption{Overall caption for the three figures showing classification results on different datasets.}
    \label{fig:all_datasets_cls}
\end{figure*}

\subsection{Classification Performance Within and Across Languages}
To evaluate practical utility and cross-lingual transferability, XTRA’s document–topic distributions \(\theta\) are used as features for downstream classification with Support Vector Machines (SVMs), following \citet{infoctm}. Performance is assessed in both intralingual (-I) and crosslingual (-C) settings. Methods from \citet{refining} (u-SVD and SVD-LR) do not produce document–topic posteriors and therefore cannot be applied to these downstream tasks, so they are omitted here. As shown in Figure~\ref{fig:all_datasets_cls}, XTRA consistently and significantly outperforms baseline models across all datasets and evaluation scenarios. For instance, on the EC News dataset, XTRA demonstrates markedly superior accuracies in both intralingual (EN-I, ZH-I) and crosslingual tasks (EN-C, ZH-C) when compared to prominent baselines. This trend of clear outperformance is consistently replicated across the Amazon Review and Rakuten Amazon datasets, where XTRA again exhibits visibly higher classification accuracies in all tested intralingual and crosslingual settings depicted. This robust and generalizable classification advantage directly stems from the superior $\theta$ representations learned by XTRA, which are effectively shaped by its integrated loss functions ($L_{\text{InfoNCE}}$, $L_{\text{Cluster}}$, $L_{\beta}$) to ensure robust semantic alignment and high discriminative capability for the SVMs.

\subsection{Ablation Study} \label{subsec:ablation}

\begin{table}[t]
\centering
\small
\setlength{\tabcolsep}{3pt}
\begin{tabular}{l c c c c c c}
\toprule
& \multicolumn{2}{c}{Topic Quality} & \multicolumn{4}{c}{Classification} \\
\cmidrule(lr){2-3} \cmidrule(lr){4-7}
Model                          & CNPMI & TU    & EN-I  & ZH-I  & EN-C  & ZH-C  \\
\midrule
NMTM$^{\dagger}$       & 0.031 & 0.784 &0.771 & 0.731 & 0.487 & 0.420 \\ 
InfoCTM$^{\dagger}$    & 0.048 & 0.913 & 0.760 & 0.747 & 0.545 & 0.556 \\ 
\midrule
w/o $L_{\beta}$             & 0.064 & 0.978 & 0.803 & \textbf{0.791} & 0.550 & 0.562 \\ 
w/o $L_{\text{cluster}}$    & 0.054 & 0.991 & 0.814 & {0.789} & \textbf{0.577} & 0.580 \\ 
w/o $L_{\text{InfoNCE}}$    & 0.051 & 0.972 & 0.785 & 0.765 & 0.372 & 0.453 \\ 
\midrule
\textbf{XTRA}                    & \textbf{0.076} & \textbf{0.993} & \textbf{0.815} & 0.788 & 0.575 & \textbf{0.582} \\
\bottomrule
\end{tabular}
\caption{Ablation study results on the ECNews dataset (50 topics). Performance is measured by Topic Quality (CNPMI, TU) and Classification Accuracy. The best results for XTRA and its ablated versions are in bold.  †Results reported in \cite{infoctm}.}
\label{tab:ablation_ecnews}
\end{table}

An ablation study on ECNews with 50 topics (Table~\ref{tab:ablation_ecnews}) confirmed the distinct contributions of XTRA’s core loss functions; the full model delivered superior topic quality, with CNPMI \(0.076\) and TU \(0.993\).  While TU remained high (above 0.97) across configurations, optimal CNPMI, crucial for cross-lingual coherence, relied on all components. Removing $L_{\text{InfoNCE}}$ reduced CNPMI to 0.051, impacting the intended enhancement of document-topic ($\theta$) alignment via PLM embeddings. Omitting $L_{\text{cluster}}$ yielded a CNPMI of 0.054, indicating less effective direct shaping of the $\theta$ space for cross-lingual consistency. The absence of $L_{\beta}$ resulted in a CNPMI of 0.064, suggesting a weakened semantic correspondence for topic-word ($\beta$) distributions. These ablated CNPMI scores, though lower than the full model, generally surpassed baselines like NMTM (0.031) and InfoCTM (0.048). For classification, XTRA was competitive, with an EN-I score of 0.815 and a ZH-C score of 0.582. The essential role of $L_{\text{InfoNCE}}$ in fostering transferable representations was clear, as its removal significantly dropped EN-C performance from 0.575 to 0.372. While configurations lacking $L_{\beta}$ (ZH-I: 0.791) or $L_{\text{cluster}}$ (EN-C: 0.577) showed specific sub-task strengths, their primary design for topic and document-topic alignment for coherence was validated as key. These findings underscore the complementary roles of each loss in XTRA's overall performance.

\begin{figure*}[!t]
    \centering
    \begin{subfigure}[b]{0.95\linewidth} 
        \centering
        \includegraphics[width=\linewidth]{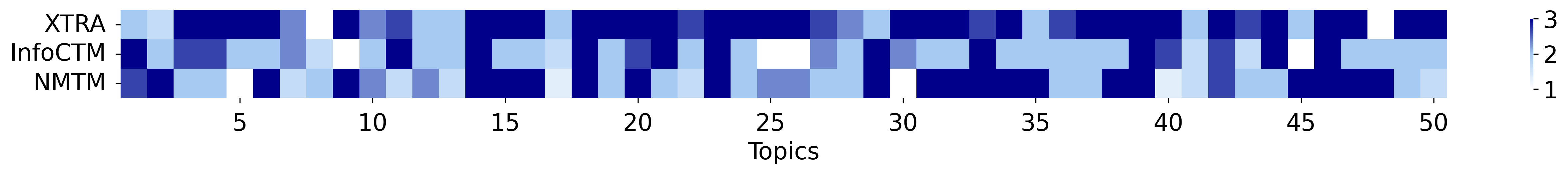} 
        \caption{English dataset (Intralingual coherence).} 
        \label{fig:A1}
    \end{subfigure}
    \vspace{1em} 

    \begin{subfigure}[b]{0.95\linewidth}
        \centering
        \includegraphics[width=\linewidth]{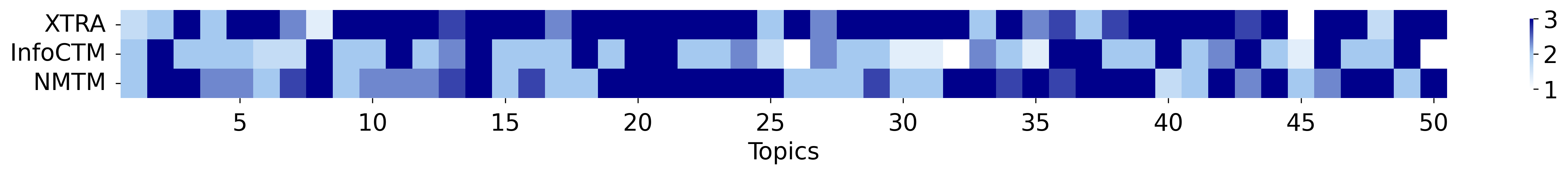} 
        \caption{Chinese dataset (Intralingual coherence).} 
        \label{fig:A2}
    \end{subfigure}
    \vspace{1em}

    \begin{subfigure}[b]{0.95\linewidth}
        \centering
        \includegraphics[width=\linewidth]{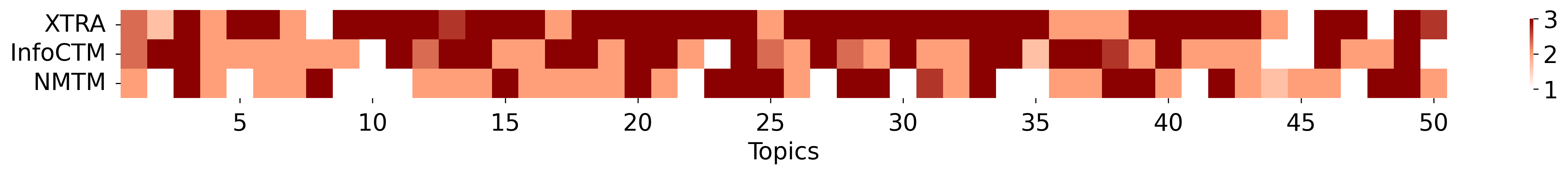} 
        \caption{Crosslingual similarity (English-Chinese).} 
        \label{fig:A}
    \end{subfigure}
    
    \caption{LLM-based topic quality evaluations on the Amazon Review dataset. Darker shades indicate higher scores. The final reported score for each evaluation is a real number, representing the average of three independent LLM assessments.}
    \label{fig:all_llm_evals} 
\end{figure*}

\subsection{Qualitative Analysis: Discovered Topic Word Examples}
\label{sec:qualitative_topic_analysis_appendix}

To further assess topic quality, Table~\ref{tab:top_related_words_new_appendix} presents the top keywords on EC News produced by the autoencoding topic models NMTM, InfoCTM, and XTRA, with noisy or misaligned terms highlighted in red.
XTRA consistently generates more coherent and better-aligned cross-lingual topics than the baselines.

\begin{CJK}{UTF8}{gbsn}
\begin{table}[!h] 
\centering
\resizebox{\linewidth}{!}{%
\begin{tabular}{@{}llccccc@{}}
    \toprule
    \multicolumn{7}{c}{\textbf{Topic: Social Media and Internet}} \\ 
    \midrule
    \multicolumn{7}{c}{NMTM Topic 45} \\
    \midrule
    \textbf{EN Topic:} & & site & online & internet & \textcolor{red}{sandy} & \textcolor{red}{allegedly} \\ 
    \textbf{ZH Topic:} & & 游戏 & 网民 & 软件 & 网页 & 网游 \\ 
    \textbf{Translations:} & & game & netizen & software & webpage & online game \\
    \midrule
    \multicolumn{7}{c}{InfoCTM Topic 32} \\ 
    \midrule
    \textbf{EN Topic:} & & offer & data & web & \textcolor{red}{ram} & remote \\ 
    \textbf{ZH Topic:} & & 网址 & 管理 & 销售 & 客户 & 互联网 \\ 
    \textbf{Translations:} & & website & manage & sales & customer & internet \\
    \midrule
    \multicolumn{7}{c}{XTRA Topic 32} \\ 
    \midrule
    \textbf{EN Topic:} & & followers & posts & posted & tweets & social \\ 
    \textbf{ZH Topic:} & & 网民 & 微博 & 转发 & 网 & 网站 \\ 
    \textbf{Topic Translations:} & & netizen & weibo & retweet & network & website \\
    \midrule\midrule
     \multicolumn{7}{c}{\textbf{Topic: Finance}} \\ 
    \midrule
     \multicolumn{7}{c}{NMTM Topic 8} \\
    \midrule
     \textbf{EN Topic:} & & stock & fund & market & \textcolor{red}{month} & \textcolor{red}{jackpot} \\ 
     \textbf{ZH Topic:} & & 基金 & 股型 & 上证 & 华夏 & 净值 \\ 
     \textbf{Translations:} & & fund & stock type & shanghai idx & huaxia & net value \\
    \midrule
    \multicolumn{7}{c}{InfoCTM Topic 15} \\
    \midrule
    \textbf{EN Topic:} & & stock & bank & share & report & due \\ 
    \textbf{ZH Topic:} & & 投资者 & \textcolor{red}{访问} & \textcolor{red}{明日} & \textcolor{red}{无尽} & 股市 \\ 
     \textbf{Translations:} & & investor & \textcolor{red}{visit} & \textcolor{red}{tomorrow} & \textcolor{red}{endless} & stock market \\
    \midrule
    \multicolumn{7}{c}{XTRA Topic 1} \\
    \midrule
    \textbf{EN Topic:} & & lenders & banks & financial & banking & debt \\ 
    \textbf{ZH Topic:} & & 信贷 & 利率 & 存款 & 银行 & 贷款 \\ 
     \textbf{Translations:} & & credit & interest rate & deposit & bank & loan \\
    \bottomrule
\end{tabular}%

}

\caption{Top-5 topic words from EC News for two selected topics ("Social Media and Internet" and "Finance") learned by NMTM, InfoCTM, and XTRA. \textcolor{red}{Red} words indicate noisy or disaligned terms with the corresponding cross-lingual topic.}
\label{tab:top_related_words_new_appendix}
\end{table}
For the Social Media and Internet topic, NMTM and InfoCTM include noisy terms (such as NMTM's ``sandy'' and ``allegedly'', or InfoCTM's ``ram'') and exhibit weak cross-lingual alignment. In stark contrast, XTRA produces a highly consistent topic. It effectively captures core concepts through semantically aligned English keywords (including ``followers'', ``posts'', ``tweets'') and their Chinese counterparts (namely ``网民 (netizen)'', ``微博 (weibo)'', ``转发 (retweet)''), all without the noise seen in other models.

A similar advantage for XTRA is evident in the Finance topic. While NMTM again introduces outliers like ``month" and ``jackpot", and InfoCTM includes noisy or off-topic words, such as ``访问 (visit)", ``明日 (tomorrow)", ``无尽 (endless)", XTRA delivers a semantically focused and cross-lingually coherent topic. It covers essential financial concepts, featuring relevant English keywords (such as ``banks", ``financial") and coherent Chinese terms (such as ``银行 (bank)", ``存款 (deposit)", "贷款 (loan)"), with no apparent noise, unlike the baselines which struggle with coherence and alignment.

\end{CJK}

\subsection{LLM-based Topic Quality Evaluation}
Inspired by recent work using large language models for automated topic model assessment \cite{llm_eval}, we incorporated LLM-based evaluations on three VAE-based topic models: NMTM, InfoCTM, and XTRA. We used LLMs for two tasks: assessing intra-lingual coherence (relatedness, 1–3 scale) and evaluating cross-lingual semantic similarity (similarity, 1–3 scale). System prompts are in Appendix~\ref{sec:llm_prompts}.

A comprehensive overview of LLM ratings for Amazon Review is presented in Figures \ref{fig:A1} (English Intralingual), \ref{fig:A2} (Chinese Intralingual), and \ref{fig:A} (Crosslingual Similarity), where darker shades indicate higher scores. These visualizations collectively indicate XTRA's strong performance. In both intralingual coherence evaluations (Figures \ref{fig:A1}, \ref{fig:A2}), XTRA consistently achieves a high concentration of top scores, performing competitively with or surpassing baselines including InfoCTM \cite{infoctm} and NMTM \cite{WuLZM20}. While models like NMTM can exhibit good top word coherence, this may associate with reduced topic diversity \cite{infoctm}. XTRA, in contrast, demonstrates strong intralingual coherence while also maintaining high topic diversity, suggesting a more balanced, robust topic generation. XTRA's benefits become particularly evident in crosslingual similarity assessment (Figure \ref{fig:A}). Here, XTRA maintains strong performance with predominantly high similarity scores, while other baselines like InfoCTM and NMTM show more varied results. This LLM-based assessment on Amazon Review suggests XTRA not only produces highly coherent topics within each language but also excels at establishing strong semantic alignment across languages, positioning it favorably against current methods.

\section{Conclusion} \label{sec:conclusion}
We propose XTRA, a cross-lingual topic modeling framework integrating Bag-of-Words with multilingual embeddings via a dual-alignment mechanism. XTRA aligns document-topic distributions through contrastive learning and projects topic-word distributions into a shared semantic space, enhancing cross-lingual consistency beyond lexical matching. Experiments show XTRA outperforms baselines in topic coherence, diversity, and alignment, demonstrating its efficacy for reliable and interpretable cross-lingual theme discovery.

\section*{Limitation}
XTRA's approach to cross-lingual topic modeling faces certain constraints. It depends on predefined numbers of topics and clusters, which limits its adaptability for datasets with ambiguous or changing themes and may result in less effective outcomes. Moreover, since it relies on pre-trained multilingual embeddings and offline clustering, its usefulness in real-time or dynamic environments is restricted, highlighting the need for further research to enhance its flexibility across various multilingual scenarios.


\section*{Ethical Considerations}
We adhere to the ACL Code of Ethics and the terms
of each codebase license. Our method aims to
advance the field of topic modeling, and we are
confident that, when used properly and with care,
it poses no significant social risks.

\section*{Acknowledgements}

Trung Le was partly supported by the Air Force Office of Scientific Research under award number FA2386-23-1-4044. 

\bibliography{acl_latex}

\appendix

\newpage
\section{Related Work}

\textbf{Topic Models and Cross-lingual topic model} Topic modeling is a core technique for discovering latent semantic structures in text corpora. The classical Latent Dirichlet Allocation (LDA) \cite{blei2003lda} models documents as mixtures of latent topics. Recent research has significantly advanced this field, integrating language representation and deep learning, leading to Neural Topic Models (NTMs). Notable NTMs include VAE-based models like Neural Variational Document Model (NVDM) and ProdLDA \cite{srivastava2017prodlda}. Advancements involve representing topics as embeddings (e.g., ETM using pre-trained embeddings \cite{dieng2020etm}, or optimal transport methods \cite{nstm,ecrtm,cite3,cite4,cite10}). Combining NTMs with pre-trained language models like BERT enhances contextual understanding \cite{BianchiTH20,BianchiTHNF21,HoyleGR20,cite5,cite6}. Contrastive learning and related regularization frameworks improve NTM training and topic–document distribution refinement \cite{2021contrastiventm,NguyenWDNNL24,cite4,cite2}, while clustering of word or document embeddings from models like Doc2Vec or BERT provides alternative topic discovery paradigms \cite{top2vec,grootendorst2022bertopic,cite5}. In addition, a substantial line of work focuses on short and noisy text, including graph convolutional models for text streams \cite{cite7}, dropout-regularized probabilistic topic models \cite{cite8}, adaptive infinite dropout for noisy and sparse data streams \cite{cite13}, infinite dropout for streaming Bayesian models \cite{cite14}, bag-of-biterms models for short texts \cite{cite15}, global clustering context methods \cite{cite5}, continual learning strategies for balancing stability and plasticity \cite{cite9}, and streaming-specific mechanisms such as out-of-vocabulary handling and topic quality control \cite{cite1}, as well as Bayesian streaming frameworks that preserve prior information \cite{cite11} and hierarchical extensions for short texts \cite{cite12}.

Cross-lingual topic modeling (CLTM) extends topic modeling to multilingual settings to discover aligned themes across languages. Early methods like \cite{MimnoWNSM09} relied on parallel corpora, limiting applicability. Later approaches utilized bilingual dictionaries for vocabulary alignment \cite{JagarlamudiD10,abs-1205-2657}, with dictionary-based translation improvements by \citet{DBLP:conf/acl/ShiLBX16}, \citet{YuanDY18}, \citet{YangBR19}, \citet{WuLZM20} and \citet{infoctm}. An alternative line uses multilingual word embeddings \cite{ChangH21}, facing issues like isomorphism assumptions. Transformer-based methods \cite{BianchiTHNF21, MuellerD21} enable zero-shot inference but still struggle with cross-lingual alignment.

\textbf{Mutual Information Maximization} 
Mutual Information Maximization (MIM) is a principle in unsupervised/self-supervised learning, often approximated by InfoNCE \cite{oord2019infonce}. It has been applied to sentence embedding \cite{simcse} and multilingual representation alignment \cite{infoxlm}. In neural topic modeling, contrastive learning based on InfoNCE has been used for discriminative topic distributions \cite{2021contrastiventm}, document-topic alignment \cite{pham2024neuromaxenhancingneuraltopic}, and extended to the cross-lingual setting for aligned topics in InfoCTM \cite{infoctm}. These works highlight MIM's effectiveness in capturing structured information.

\newpage
\section{Algorithm}
In this section, we present the \textbf{XTRA} training procedure: 
\begin{algorithm}[H]
\caption{XTRA training procedure}
\label{alg:xtra_glocom_style_condensed}
\begin{algorithmic}[1]
    \REQUIRE Input corpus $\mathbf{X} = \mathbf{X}^{(1)} \cup \mathbf{X}^{(2)}$, $K$, $N$, $C$, $\lambda_{1,2,3}$.
    \ENSURE Optimized parameters $\Theta^* = \{\text{Encoders}^*, \beta^{(1)*}, \beta^{(2)*}, \text{Projectors}^*\}$
    \STATE Initialize parameters $\Theta = \{\text{Encoders}, \beta^{(1)}, \beta^{(2)}, \text{Projectors}\}$ and Optimizer.
    \FOR{epoch from 1 to $N$ \textbf{do}}
        \STATE Shuffle $\mathbf{X}^{(1)}$ and $\mathbf{X}^{(2)}$.
        \FOR{each balanced mini-batch $b$ sampled from $\mathbf{X}$ \textbf{do}}
            \STATE Compute document-level components: $\{\theta_d\}$, $L_{TM}^{b}$, $L_{InfoNCE}^{b}$ for all $x_d \in b$;
            \STATE Compute batch-level cluster loss $L_{Cluster}^{b}$ using $\{\theta_d\}$ and $C$;
            \STATE Compute batch-level beta alignment loss $L_{\beta}^{b}$ using $\beta^{(1)}, \beta^{(2)}, P^{1,2}$;
            \STATE $\mathcal{L}_{\text{batch}} \leftarrow L_{TM}^{b} + \lambda_1 L_{InfoNCE}^{b} + \lambda_2 L_{Cluster}^{b} + \lambda_3 L_{\beta}^{b}$;
            \STATE Update $\Theta$ with $\nabla \mathcal{L}_{\text{batch}}$;
        \ENDFOR
    \ENDFOR
\end{algorithmic}
\end{algorithm}

\section{Clustering Method}
\label{appendix:clustering}
We use an asymmetric clustering approach to ensure cross-lingual alignment.
After reducing dimensionality via Singular Value Decomposition (SVD)~\cite{deerwester1990lsi}
and applying L2 normalization, we perform KMeans clustering~\cite{macqueen1967kmeans} 
on the pivot language (e.g., English).
Documents from the other language are then assigned to the nearest clusters based on cosine similarity.
This avoids the issue of monolingual clusters that can arise when clustering both languages jointly, leading to better cross-lingual consistency.

\section{Detailed Prompts for LLM Evaluation} 
\label{sec:llm_prompts} 

This appendix provides the detailed system prompts used for the LLM-based evaluation tasks described in the main text. Tables~\ref{tab:intralingual_prompts} and \ref{tab:crosslingual_prompts} shows the prompts side-by-side for intralingual coherence and crosslingual similarity assessment across the different datasets.

\begin{table*}[htbp]
\centering
\begin{tabular}{|>{\centering\arraybackslash}p{0.1\textwidth}|p{0.85\textwidth}|}
\hline
\rowcolor{gray!10}
\textbf{Dataset} & \multicolumn{1}{c|}{\textbf{Prompt}} \\
\hline

\multirow{6}{*}{\parbox[c]{0.1\textwidth}{\centering\textbf{EC\\[0.5ex] News}}} & 
You are a helpful assistant evaluating the top words of a topic model output for a given topic. The dataset is EC News, a collection of English and Chinese news with 6 categories: business, education, entertainment, sports, tech, and fashion. Please rate how related the following words are to each other on a scale from 1 to 3 ("1"=not very related, "2"=moderately related, "3"=very related). Reply with a single number, indicating the overall appropriateness of the topic. \\
\hline

\multirow{7}{*}{\parbox[c]{0.1\textwidth}{\centering\textbf{Amazon Review}}} & 
You are a helpful assistant evaluating the top words of a topic model output for a given topic. The dataset is Amazon Review, which includes English and Chinese reviews from the Amazon website. Please rate how related the following words are to each other on a scale from 1 to 3 ("1"=not very related, "2"=moderately related, "3"=very related). Reply with a single number, indicating the overall appropriateness of the topic. \\
\hline

\multirow{6}{*}{\parbox[c]{0.1\textwidth}{\centering\textbf{Rakuten Amazon}}} & 
You are a helpful assistant evaluating the top words of a topic model output for a given topic. The dataset is Rakuten Amazon, which contains Japanese reviews from Rakuten, and English reviews from Amazon. Please rate how related the following words are to each other on a scale from 1 to 3 ("1"=not very related, "2"=moderately related, "3"=very related). Reply with a single number, indicating the overall appropriateness of the topic. \\
\hline

\end{tabular}
\caption{Intralingual Coherence Prompts for LLM-based Evaluation}
\label{tab:intralingual_prompts}

\end{table*}

\begin{table*}[htbp]
\centering

\begin{tabular}{|>{\centering\arraybackslash}p{0.1\textwidth}|p{0.85\textwidth}|}
\hline
\rowcolor{gray!10}
\textbf{Dataset} & \multicolumn{1}{c|}{\textbf{Prompt}} \\
\hline

\multirow{6}{*}{\parbox[c]{0.1\textwidth}{\centering\textbf{EC\\[0.5ex] News}}} & 
You are a helpful assistant evaluating the similarity of topics derived from topic modeling on parallel news corpora. The dataset is EC News, with English and Chinese news. You will be given two sets of top words, one for an English topic (Language 1) and one for a Chinese topic (Language 2). Please rate how similar the underlying topics represented by these two sets of words are, on a scale from 1 to 3 ("1"=not very similar, "2"=moderately similar, "3"=very similar). Reply with a single number. \\
\hline

\multirow{6}{*}{\parbox[c]{0.1\textwidth}{\centering\textbf{Amazon Review}}} & 
You are a helpful assistant evaluating the similarity of topics derived from topic modeling on parallel review corpora. The dataset is Amazon Review, with English and Chinese reviews. You will be given two sets of top words, one for an English topic (Language 1) and one for a Chinese topic (Language 2). Please rate how similar the underlying topics represented by these two sets of words are, on a scale from 1 to 3 ("1"=not very similar, "2"=moderately similar, "3"=very similar). Reply with a single number. \\
\hline

\multirow{7}{*}{\parbox[c]{0.1\textwidth}{\centering\textbf{Rakuten Amazon}}} & 
You are a helpful assistant evaluating the similarity of topics derived from topic modeling on parallel review corpora. The dataset is Rakuten Amazon, with Japanese reviews (Rakuten - Language 2) and English reviews (Amazon - Language 1). You will be given two sets of top words, one for an English topic and one for a Japanese topic. Please rate how similar the underlying topics represented by these two sets of words are, on a scale from 1 to 3 ("1"=not very similar, "2"=moderately similar, "3"=very similar). Reply with a single number. \\
\hline

\end{tabular}
\caption{Crosslingual Similarity Prompts for LLM-based Evaluation}
\label{tab:crosslingual_prompts}

\end{table*}

\section{Implementation Details}
Our models were trained on a single NVIDIA P100 GPU (Kaggle). We employed the Adam optimizer~\cite{kingma2014adam} with an initial learning rate of 0.002 and trained for 800 epochs. A learning rate scheduler decayed the learning rate by a factor of 0.5 every 250 epochs. Hyperparameters were tuned over:
\begin{itemize}
    \item \(\lambda_1\) for \(\mathcal{L}_{\text{InfoNCE}}\): \{70, 80, 85\}
    \item \(\lambda_2\) for \(\mathcal{L}_{\text{Cluster}}\): \{5, 10\}
    \item \(\lambda_3\) for \(\mathcal{L}_{\beta}\): \{7, 15\}
\end{itemize}
We report results with the best hyperparameters. On Amazon Reviews, XTRA trained in about 30 minutes on a single P100; InfoCTM took about 35 minutes; NMTM finished in about 7 minutes and 35 seconds. Rakuten Amazon showed comparable wall-clock times under the same hardware and settings. On ECNews, training ran longer: around 60 minutes for XTRA, 70 for InfoCTM, and 14 for NMTM.

    \end{document}